\documentclass[conference]{IEEEtran}
\usepackage[a4paper, top=19mm, bottom=43mm, left=14mm, right=14mm]{geometry}
\usepackage{cite}
\usepackage{amsmath,amssymb,amsfonts}
\usepackage{algorithmic}
\usepackage{graphicx}
\usepackage{textcomp}
\usepackage{xcolor}
\usepackage[hang,small,bf]{caption}
\usepackage[subrefformat=parens]{subcaption}
\def\BibTeX{{\rm B\kern-.05em{\sc i\kern-.025em b}\kern-.08em
    T\kern-.1667em\lower.7ex\hbox{E}\kern-.125emX}}
\begin{document}

\title{Privacy-Preserving Vision Transformer Using Images Encrypted with Restricted Random Permutation Matrices}

\author{\IEEEauthorblockN{1\textsuperscript{st} Kouki Horio}
\IEEEauthorblockA{\textit{Tokyo Metropolitan University} \\
6-6, Asahigaoka, Hino, Tokyo, Japan \\
horio-kouki@ed.tmu.ac.jp}
\and
\IEEEauthorblockN{2\textsuperscript{nd} Kiyoshi Nishikawa}
\IEEEauthorblockA{\textit{Tokyo Metropolitan University} \\
6-6, Asahigaoka, Hino, Tokyo, Japan \\
kiyoshi@tmu.ac.jp}
\and
\IEEEauthorblockN{3\textsuperscript{rd} Hitoshi Kiya}
\IEEEauthorblockA{\textit{Tokyo Metropolitan University} \\
6-6, Asahigaoka, Hino, Tokyo, Japan \\
kiya@tmu.ac.jp}
}

\maketitle

\begin{abstract}
We propose a novel method for privacy-preserving fine-tuning vision transformers (ViTs) with encrypted images. Conventional methods using encrypted images degrade model performance compared with that of using plain images due to the influence of image encryption. In contrast, the proposed encryption method using restricted random permutation matrices can provide a higher performance than the conventional ones.
\end{abstract}


\section{Introduction}
The use of cloud environments has been increasing in various applications of deep neural networks (DNNs). However, cloud environments are not always trusted in general, so privacy-preserving deep learning has become an urgent problem\cite{SIP-2021-0048, maung_privacy, 9802995, 9934926, 10483907, 8486525}.

One of the privacy-preserving solutions for DNNs is to use images encrypted with learnable encryption to protect sensitive visual information in images for training and testing models. However, in this approach, most existing methods using encrypted images have a problem, that is, the performance of models degrades compared with models without encryption\cite{iijima2024, madono2020, SIA-GAN}.  

Accordingly, in this paper, we propose a novel method using encrypted images that can reduce the performance degradation of models under the use of the vision transformer (ViT)\cite{ViT}. In the proposed method, a pre-trained model is fine-tuned by using images encrypted with restricted random matrices generated with secret keys, and query images encrypted with the same keys are applied to the fine-tuned model. In experiments, it is verified that the method can provide a higher performance than conventional ones.

\section{Proposed method}

\subsection{Overview of proposed method}
\begin{figure}[bth]
    \centering
    \includegraphics[bb=0 0 806 500, scale=0.29]{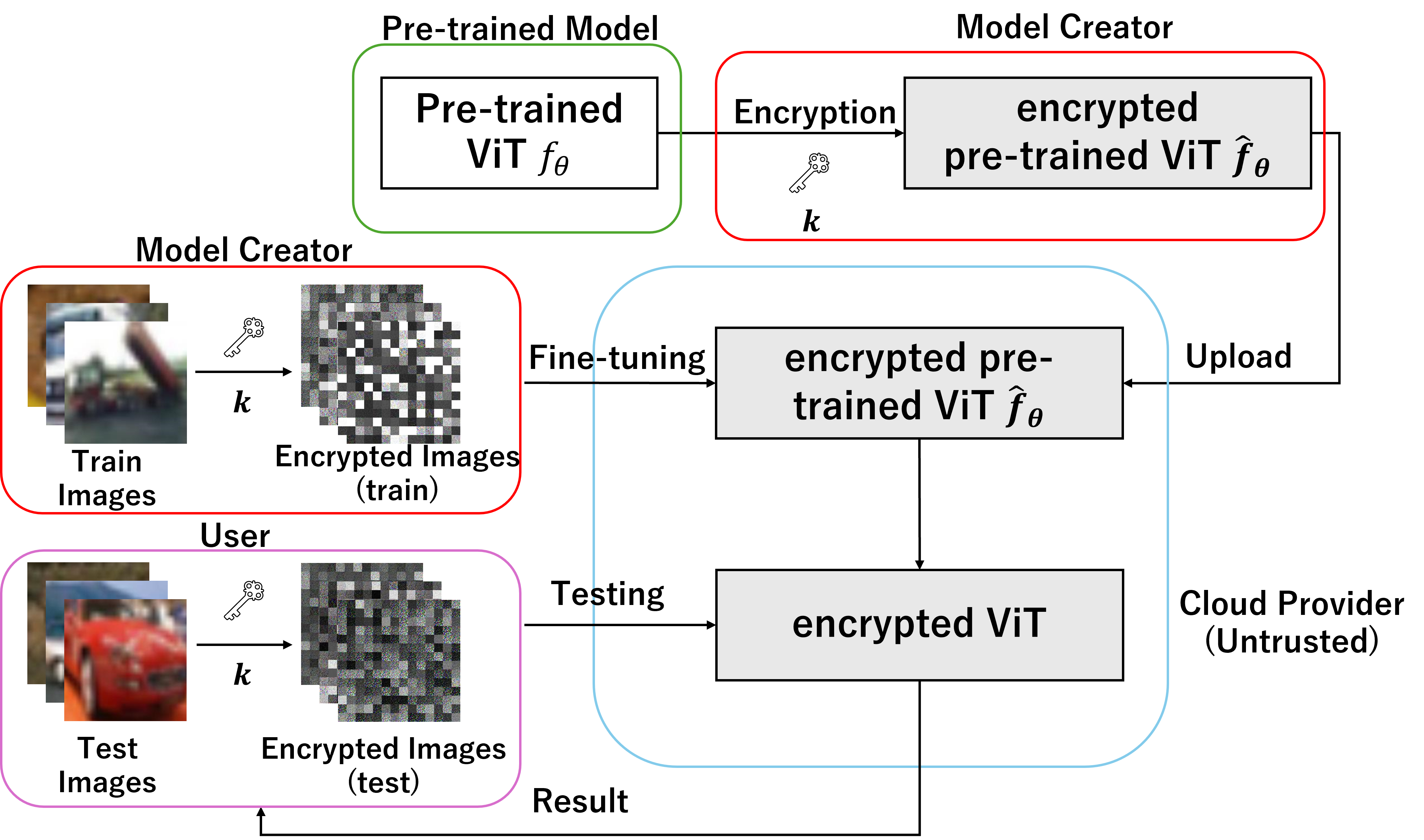}
    \caption{Framework of privacy-preserving ViT}
\end{figure}

Figure 1 shows an overview of privacy-preserving ViTs using encrypted images in an image classification task where the cloud provider is assumed to be untrusted. A model creator encrypts training images with secret keys and send a pre-trained model and the encrypted images to a cloud provider. The provider fine-tunes the pre-trained model by using the encrypted images. In contrast, a client encrypts a query image with the same key as the keys used for encrypting training images and send it to the provider. The provider classifies the encrypted query with the trained model, and provide an estimation result to the client. Note that the provider has not both visual information of plain images and secret keys.

\subsection{Image encryption}
Fine-tuning and testing are carried out by using encrypted images as shown in Fig. 1. ViT has a high similarity to block-wise encryption\cite{iijima2024, nagamori2024efficient, KIYA20232022}, so the proposed method is carried out using block-wise encryption, which consists of block scrambling (block permutation) and pixel permutation.

Below is the procedure for image encryption, in which $\boldsymbol{\rm{E_{bs}}}$ is a random permutation matrix for block permutation, given by
\small
\begin{equation}
    \label{Ebs}
    \boldsymbol{\rm{E_{bs}}} = 
        \begin{pmatrix}
            E_{bs}(1,1) & \dots & E_{bs}(1,j) & \dots & E_{bs}(1,N) \\
            \vdots &  & \vdots &  & \vdots \\
            E_{bs}(i,1) & \dots & E_{bs}(i,j) & \dots & E_{bs}(i,N) \\
            \vdots &  & \vdots &  & \vdots \\
            E_{bs}(N,1) & \dots & E_{bs}(N,j) & \dots & E_{bs}(N,N) \\
        \end{pmatrix}
    .
\end{equation}
\normalsize

\begin{itemize}
    \item [1)] Divide an image $x \in {\mathbb{R}^{h \times w \times c}}$ into $N$ non-overlapped blocks with a size of $p \times p$ such that $B = {\{B_1, \cdots ,B_i, \cdots, B_N\}}^{\top}$, $B_i \in {\mathbb{R}^{p^2c}}$, where $h, w$ and $c$ are the height, width, and number of channels in the image, and $p$ is also equal to the patch size of ViT.

    \item [2)] Using key $k_{1}$, generate a random integer vector with a length of $N$ as
    \begin{align}
        l = [l(1), \cdots, l(i), \cdots, l(N)]
    \end{align}
    where,
    \begin{gather}
        l(i)\in\{1, 2, \cdots, N\}, \nonumber \\
        l(i)\ne l(j) \quad
        \text{if} \quad i \ne j, \quad
        i,j\in\{1, 2, \cdots, N\}. \nonumber
    \end{gather}
    
    \item [3)] Define $\boldsymbol{\rm{E_{bs}}}$ in Eq.(\ref{Ebs}) as
    \begin{equation}
        E_{bs}(i,j) = 
        \begin{cases}
            0 & \text{if $l(j) \ne i$,}\\
            1 & \text{if $l(j) = i$}
        \end{cases}
        .
    \end{equation}
    
    \item [4)] Define a vector $s$ as
    \begin{equation}
        s = [1, 2,\cdots, N],
    \end{equation}
    and transform it with $\boldsymbol{\rm{E_{bs}}}$ as
    \begin{align}
        \hat{s} &= s\boldsymbol{\rm{E_{bs}}} \\
        &= [\hat{s}(1), \hat{s}(2),\cdots, \hat{s}(N)], \hat{s}(i) \in \{1, 2,\cdots, N\}. \nonumber
    \end{align}
    
    \item [5)] Give permuted blocks $\hat{B} = {\{\hat{B_1}, \cdots ,\hat{B_i}, \cdots, \hat{B_N}\}}^{\top}$ as
    \begin{equation}
        \hat{B}_i = B_{\hat{s}(i)}.
    \end{equation}
\end{itemize}

Next, the procedure for pixel permutation is explained. In the method, $R, G,$ and $B$ values in each block are independently permuted. A random permutation matrix for pixel permutation $\boldsymbol{\rm{E_{ps}}}$, which is applied to each block, is expressed as
\small
\begin{equation}
    \label{Eps}
    \boldsymbol{\rm{E_{ps}}} = 
        \begin{pmatrix}
            E_{ps}(1,1) & \dots & E_{ps}(1,j) & \dots & E_{ps}(1,L) \\
            \vdots &  & \vdots &  & \vdots \\
            E_{ps}(i,1) & \dots & E_{ps}(i,j) & \dots & E_{ps}(i,L) \\
            \vdots &  & \vdots &  & \vdots \\
            E_{ps}(L,1) & \dots & E_{ps}(L,j) & \dots & E_{ps}(L,L) \\
        \end{pmatrix}
    .
\end{equation}
\normalsize
where $L = p^2c$.

Below is the procedure for the permutation.
\begin{itemize}
    \item [6)] Using key $k_{2}$, generate a random integer vector with a length of $L$ as
    \begin{align}
        u = [u(1), \cdots, u(i), \cdots, u(L)]
    \end{align}
    where,
    \begin{gather}
        u(i)\in\{1, 2, \cdots, L\}, \nonumber \\
        u(i)\ne u(j) \quad
        \text{if} \quad i \ne j, \quad
        i,j\in\{1, 2, \cdots, L\}. \nonumber
    \end{gather}
    
    \item [7)] Define $\boldsymbol{\rm{E_{ps}}}$ in Eq.(\ref{Eps}) as
    \begin{equation}
        E_{ps}(i,j) = 
        \begin{cases}
            0 & \text{if $u(j) \ne i$,}\\
            1 & \text{if $u(j) = i$}
        \end{cases}
        .
    \end{equation}
    
    \item [8)] Vectorize the elements of each block $\hat{B}_i$ as $\hat{b}_i \in \mathbb{R}^{L}$, and transform it with $\boldsymbol{\rm{E_{ps}}}$ with
    \begin{equation}
        b^{'}_i = \hat{b}_i\boldsymbol{\rm{E_{ps}}}.
    \end{equation}
    
    \item [9)] Concatenate the encrypted vectors $b_i^{\prime}$ into an encrypted image $x^{\prime}$.
\end{itemize}

\subsection{Example of restricted random permutation matrices}
A permutation matrix is a square binary matrix that has exactly one entry of 1 in each row and each column with all other entries 0. Every permutation matrix P is orthogonal, with its inverse equal to its transpose.

Three types of permutation matrices are considered. For $L = 5$, an example of $\boldsymbol{\rm{E_{ps}}}$ is given below where * indicates a fixed element.

\begin{itemize}
    \item[A)] Identity matrix ($N_{ps} = L$): \\
    \begin{equation}
        \label{A)}
        \boldsymbol{\rm{E_{ps}}} = 
            \begin{pmatrix}
                1^* & 0 & 0 & 0 & 0 \\
                0 & 1^* & 0 & 0 & 0 \\
                0 & 0 & 1^* & 0 & 0 \\
                0 & 0 & 0 & 1^* & 0 \\
                0 & 0 & 0 & 0 & 1^*
            \end{pmatrix}
        .
    \end{equation}
    If $\boldsymbol{\rm{E_{ps}}}$ is the identity matrix of $L \times L$, in which all its diagonal elements equal 1, and 0 everywhere else, no permutation is carried out. In this type, the number of fixed diagonal elements $N_{ps}$ is given as $N_{ps} = L$.
    
    \item[B)] Restricted random permutation matrix ($0 < N_{ps} < L$): \\
    $N_{ps} < L$ diagonal elements are fixed as a value of 1, where the positions of the elements are randomly selected. When using $0 < N_{ps} < L$, a permutation matrix is called a restricted random permutation matrix. For $N_{ps} = 2$,
    \begin{equation}
        \label{B)}
        \boldsymbol{\rm{E_{ps}}} = 
            \begin{pmatrix}
                1^* & 0 & 0 & 0 & 0 \\
                0 & 0 & 0 & 0 & 1 \\
                0 & 0 & 1* & 0 & 0 \\
                0 & 1 & 0 & 0 & 0 \\
                0 & 0 & 0 & 1 & 0
            \end{pmatrix}
        .
    \end{equation}
    
    \item[C)] Unrestricted random permutation matrix ($N_{ps} = 0$): \\
    The positions of $L$ elements are randomly selected as
    \begin{equation}
        \label{C)}
        \boldsymbol{\rm{E_{ps}}} = 
            \begin{pmatrix}
                0 & 0 & 0 & 1 & 0 \\
                0 & 0 & 1 & 0 & 0 \\
                0 & 0 & 0 & 0 & 1 \\
                1 & 0 & 0 & 0 & 0 \\
                0 & 1 & 0 & 0 & 0
            \end{pmatrix}
        .
    \end{equation}
\end{itemize}

The above equations correspond to a pixel shuffling (permutation) operation in each patch. For $N_{ps} = L$, Eq.(\ref{B)}) is reduced to as Eq.(\ref{A)}), and Eq.(\ref{B)}) is reduced to as the type of C) for $N_{ps} = 0$. Similarly, $\boldsymbol{\rm{E}_{bs}}$, which is also a permutation matrix, is classified into three types, so $N_{bs} < N$ diagonal elements are fixed as in eq.(\ref{B)}). The transformation with $\boldsymbol{\rm{E}_{bs}}$ is called block scrambling or block permutation. Fig. 2 shows an example of encrypted images where $N=196$ and $L=768$ were used.

\begin{figure}[h]
    \centering
    \begin{minipage}[b]{0.48\linewidth}
        \centering
        \includegraphics[bb= 0 0 175 170, scale=0.6]{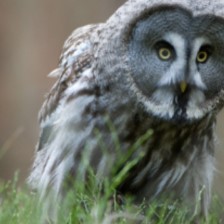}
        \subcaption{Plain}
    \end{minipage}
    \begin{minipage}[b]{0.48\linewidth}
        \centering
        \includegraphics[bb= 0 0 170 170, scale=0.6]{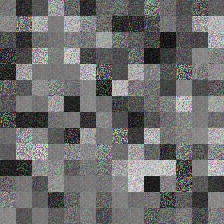}
        \subcaption{$N_{bs}=0, N_{ps}=0$}
    \end{minipage} \\

    \begin{minipage}[b]{0.48\linewidth}
        \centering
        \includegraphics[bb= 0 0 175 180, scale=0.6]{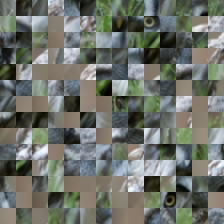}
        \subcaption{$N_{bs}=0, N_{ps}=768$}
    \end{minipage}
    \begin{minipage}[b]{0.48\linewidth}
        \centering
        \includegraphics[bb= 0 0 170 180, scale=0.6]{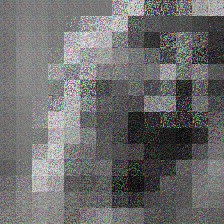}
        \subcaption{$N_{bs}=196, N_{ps}=0$}
    \end{minipage} \\

    \begin{minipage}[b]{0.48\linewidth}
        \centering
        \includegraphics[bb= 0 0 175 180, scale=0.6]{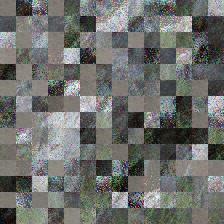}
        \subcaption{$N_{bs}=60, N_{ps}=350$}
    \end{minipage}
    \begin{minipage}[b]{0.48\linewidth}
        \centering
        \includegraphics[bb= 0 0 170 180, scale=0.6]{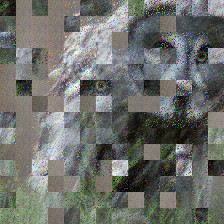}
        \subcaption{$N_{bs}=120, N_{ps}=500$}
    \end{minipage}
    \caption{Example of encrypted images}
\end{figure}

\section{Experiments}
The effectiveness of the proposed method is demonstrated in the efficiency of model training and the accuracy of classifications on the CIFAR-10 dataset. In experiments, a pre-trained model 'vit\_base\_patch16\_224' from the timm library was used. Fine-tuning was carried out under the use of a batch size of 64, a learning rate of 0.0001, a momentum of 0.9, and a weight decay of 0.0005 using the stochastic gradient descent (SGD) algorithm for 15 epochs. A cross-entropy loss function was used as the loss function. $N=196$ and $L=768$ were used.

\subsection{Classification accuracy}
In table.1, the classification accuracy and loss value of each model per epoch during testing are evaluated. From the table, when fine-tuning the pre-trained model by using encrypted images with $N_{bs}=N_{ps}=0$ (conventional), the accuracy of the fine-tuned model decreased compared with that of using plain images (Baseline). In contrast, when fine-tuning the model with larger $N_{bs}$ and $N_{ps}$ values, the accuracy was higher. Accordingly, we confirmed that the accuracy of encrypted ViTs can be managed by choosing the values of $N_{bs}$ and $N_{ps}$.

\begin{table}[h]
    \caption{Classification accuracy}
    \label{table:SpeedOfLight}
    \centering
    \begin{tabular}{c|l}
        \hline
        Encryption & Accuracy \\
        \hline \hline
        proposed ($N_{bs}=196, \quad N_{ps}=0$) & \textbf{97.49} \\
        proposed ($N_{bs}=120, \quad N_{ps}=500$) & \textbf{97.30} \\
        proposed ($N_{bs}=0, \quad N_{ps}=768$) & \textbf{88.88} \\
        proposed ($N_{bs}=60, \quad N_{ps}=350$) & \textbf{84.01} \\
        Conventional ($N_{bs}=0, \quad N_{ps}=0$) & \textbf{61.80} \\
    \hline
    Baseline ($N_{bs}=196, \quad N_{ps}=768$) & \textbf{99.05} \\
    \hline
    \end{tabular}
\end{table}

\subsection{Training efficiency}
Training efficiency is the time required for model training and loss convergence. From Fig.3, the model fine-tuned by using encrypted images with $N_{bs} = N_{ps} = 0$ (conventional) needed a larger number of epochs to achieve a high accuracy and a low loss value. In contrast, when using the model fine-tuned by using with larger $N_{bs}$ and $N_{ps}$ values, the training efficiency of the models was higher. We also observed a similar trend for validation.

\begin{figure}[htbp]
    \centering
    \begin{subfigure}[b]{0.5\textwidth}
        \centering
        \includegraphics[scale=0.35]{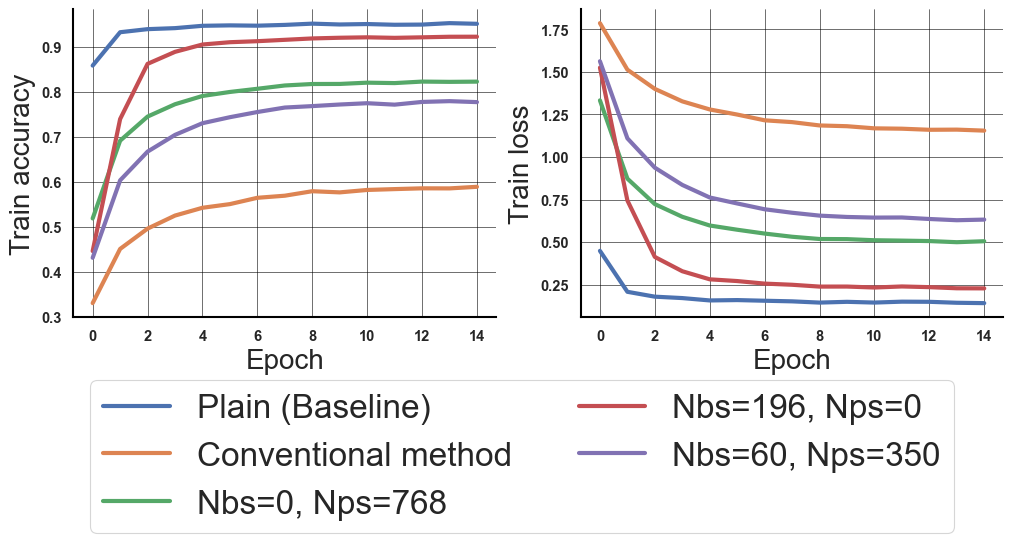}
        \subcaption{Train}
    \end{subfigure}
    \begin{subfigure}[b]{0.5\textwidth}
        \centering
        \includegraphics[scale=0.35]{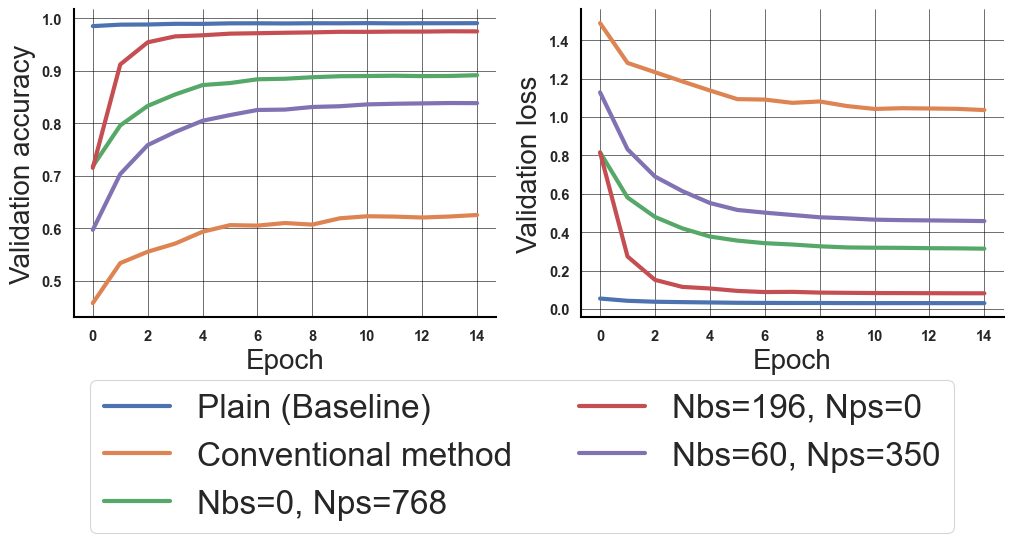}
        \subcaption{Validation}
    \end{subfigure}

    \caption{Learning curves}
    \label{fig:combined}
\end{figure}

\section{Conclusion}
In this paper, we proposed a novel method to reduce the influence of image encryption for privacy preserving ViTs. In experiments, the method was demonstrated not only to reduce the performance degradation of models but to also avoid an increase in training time even when using encrypted images.

\section*{Acknowledgement}
This work was supported in part by JSPS KAKENHI (Grant Number JP21H01327).

\bibliography{ref}
\bibliographystyle{unsrt}

\end{document}